\documentclass[conference]{IEEEtran}
\IEEEoverridecommandlockouts
\usepackage{cite}
\usepackage{amsmath,amssymb,amsfonts}
\usepackage{algorithmic, adjustbox, multirow}
\usepackage{graphicx}
\usepackage{textcomp}
\usepackage{xcolor}
\def\BibTeX{{\rm B\kern-.05em{\sc i\kern-.025em b}\kern-.08em
    T\kern-.1667em\lower.7ex\hbox{E}\kern-.125emX}}
\begin{document}

\title{Dynamic Analysis and an Eigen Initializer for Recurrent Neural Networks

\thanks{This work was partially supported by ARO grant W911NF-21-1-0254.}
}

\author{\IEEEauthorblockN{1\textsuperscript{st} Ran Dou}\\
\IEEEauthorblockA{\textit{Department of Electrical and Computer Engineering} \\
\textit{University of Florida}\\
Gainesville, FL, United States \\
dour@ufl.edu}
\and
\IEEEauthorblockN{2\textsuperscript{nd} Jose Principe}
\IEEEauthorblockA{\textit{Department of Electrical and Computer Engineering} \\
\textit{University of Florida}\\
Gainesville, FL, United States \\
principe@cnel.ufl.edu}
}

\maketitle

\begin{abstract}
In recurrent neural networks, learning long-term dependency is the main difficulty due to the vanishing and exploding gradient problem. Many researchers are dedicated to solving this issue and they proposed many algorithms. Although these algorithms have achieved great success, understanding how the information decays remains an open problem. In this paper, we study the dynamics of the hidden state in recurrent neural networks. We propose a new perspective to analyze the hidden state space based on an eigen decomposition of the weight matrix. We start the analysis by linear state space model and explain the function of preserving information in activation functions. We provide an explanation for long-term dependency based on the eigen analysis. We also point out the different behavior of eigenvalues for regression tasks and classification tasks. From the observations on well-trained recurrent neural networks, we proposed a new initialization method for recurrent neural networks, which improves consistently the performance. It can be applied to vanilla-RNN, LSTM, and GRU. 
We test is on many datasets, such as Tomita Grammars, pixel-by-pixel MNIST dataset, and machine translation dataset (Multi30k). It outperforms Xavier initializer and kaiming initializer as well as other RNN-only initializers like IRNN and sp-RNN in several tasks. 
\end{abstract}


\section{Introduction}
In a recurrent system, the hidden state is considered representative of the unknown state of the system that created the data. In general, a state space model approximates the unknown state $h_t$ using the previous hidden state $h_{t-1}$ and current sample $x_t$. And the prediction $y_t$ is observed by the hidden state. 
\begin{equation}
    h_t = H (x_t, h_{t-1})
\label{state}
\end{equation}
\begin{equation}
    y_t = G(h_t)
\end{equation}
where $H$ is the state transition function and $G$ is the observation function. Extensive research has been studied on the design of state transition functions, such as the Kalman Filter family \cite{whitney1984kalman}, echo state networks\cite{jaeger2002adaptive} and recurrent neural networks. The hidden states represent the memory of past information, however, information stored in hidden states is fading away through time. Models have difficulty learning the long-term dependencies. 

Many mechanisms have been proposed to improve long-term memory. One is to use the gating mechanism \cite{hochreiter1997long, GRU, gu2020improving} and these gated networks have been widely used in time series, reinforcement learning, and natural language processing. Generally, the output of a gated unit is limited from 0 to 1 by a sigmoid activation function. The forgetting gate is used to reset the hidden state memory and the writing gate decides what information should be stored in the hidden states. In LSTM \cite{hochreiter1997long}, the forgetting and writing gates operate a cell state vector as the memory. And the GRU operates directly on the hidden state \cite{GRU}. In order to learn the long-term dependency, a bias of 1 can be added to the forgetting gate for initialization \cite{gatebias}.

Another mechanism for improving the long-term dependency is using external memory, such as the Neural Turing Machine (NTM \cite{NTM}), Differentiable Neural Computer \cite{DNC}, End-to-end Network \cite{MemN2N}, Neural Stack and Neural Queue \cite{NeuralStack}, RNN-EM \cite{RNNEM}, and Dynamic Memory Networks \cite{DMN}. In these architectures, a recurrent network is used as the controller to interact with the external memory. At each time, the controller reads and writes from the external memory based on a rule such as the attention mechanism. Unlike overwriting new information into hidden states, the external memory stores all past information and provides a better long-term dependency.

Besides the two mechanisms mentioned above, there are other studies trying to improve the performance of RNNs. And most of them are focusing on manipulating the hidden state, such as AntisymmetricRNN \cite{chang2019antisymmetricrnn}, Noisy RNN \cite{lim2021noisy}, PF-RNN \cite{ma2020particle}, SRNN \cite{rotman2021shuffling}, TP-RNN \cite{ma2020temporal}, uRNN \cite{arjovsky2016unitary} and SR-RNN \cite{stateregularized}. These methods show that well-designed hidden states are capable of handling longer memory. Other methods are focusing more on the learning algorithms for RNNs. For example, \cite{allen2019convergence} shows that with a sufficiently large architecture, the stochastic gradient descent can provide a linear convergence rate. \cite{zhang2021sbo} proposed a stochastic bilevel optimization for RNN to prevent vanishing and exploding gradients. And \cite{massart2022coordinate} proposed to use the stochastic Riemannian coordinate descent for RNN training.

There is also research studying the effect of different initialization methods for RNNs. In gradient descent methods, a proper initialization for weights is of vital importance. If the weights are initialized with small values, the gradients will vanish and if the weights are initialized with large values, the gradients will explode. And the rules of thumb for weight initialization are to set the mean of activations close to zero and the variance of activations consistent with each other. Followed by this idea, Xavier initializer \cite{glorot2010understanding} and kaiming initializer \cite{he2015delving} are proposed and frequently used in nowadays research. Unlike feedforward neural networks, recurrent neural networks share weights recurrently, which provides a different dynamic for the hidden states. IRNN \cite{le2015simple} shows that RNNs can be simply initialized by identity matrices and \cite{talathi2015improving} proposed the np-RNN afterward. However, both methods are designed for vanilla RNNs and can not be applied in general applications such as LSTM and GRU.

In this paper, we start by analyzing a linear state space model in terms of eigendecomposition and extrapolate it to nonlinear cases. We conclude that in RNNs, the long-term dependency is improved by increasing eigenvalues. Based on our observation, we propose a new initializer for recurrent layers. Unlike IRNN and np-RNN, which are only designed for vanilla-RNN, our initializer can be widely used for different types of RNNs, like LSTM and GRU.

\section{Eigen Decomposition for State Transition Function}
\label{sec:2}
Given a linear state space model, the state transition function and observation function can be written as,
\begin{equation}
    h_t = W_h h_{t-1} + W_x x_t
\end{equation}
\begin{equation}
    y_t = W_y h_t
\end{equation}
where $h_t$ is the state vector with the size of $n$. For simplicity, denote,
\begin{equation}
    x'_t = W_x x_t
\end{equation}
For the state transition matrix $W_h$, let $\Lambda = [\lambda_1, \lambda_2, ..., \lambda_n]$ and $U = [u_1, u_2, ..., u_n]$ be the eigenvalues and eigenvectors,
\begin{equation}
    W_h = U \Lambda \bar{U}^T
\end{equation}
Note here $[u_1, u_2, ..., u_n]$ are the orthonormal basis, and $\Lambda$ and $U$ may have complex values. When the eigenvalues are real values, after each update, the hidden state is enlarged along the direction of eigenvectors by the corresponding eigenvalues. And when the eigenvalues are complex numbers, they show up in conjugate pairs and the hidden state also rotates in the plane formed by the corresponding conjugate eigenvectors.

Consider only the hidden state with zero inputs, we can decompose $h_{t-1}$ by the eigenvectors,
\begin{equation}
    h_{t-1} = \sum_{i=1}^n \alpha^i_{t-1} u_i
\end{equation}
where $[\alpha^1_{t-1}, \alpha^2_{t-1}, ..., \alpha^n_{t-1}]$ are the coefficients for the corresponding eigenvectors. And,
\begin{equation}
    h_t = \sum_{i=1}^n \alpha^i_{t} u_i= \sum_{i=1}^n \alpha^i_{t-1} \lambda_i u_i
\label{eq:h_t}
\end{equation}
After $t'$ steps,
\begin{equation}
    h_{t+t'} = \sum_{i=1}^n \alpha^i_{t-1} \lambda^{t'}_i u_i
\label{eq:decay}
\end{equation}
It is obvious that the system is stable only when $|\lambda_i| \leq 1$. And as time goes on, the information of hidden states is decaying at different speeds along different eigenvectors. 

Now take the input $x'_t$ into consideration. Same as the hidden state, we can decompose the $x'_t$ by the eigenvectors $u_i$ and its coefficients $a^i_t$.
\begin{equation}
    x'_t = \sum_{i=1}^n a^i_t u_i 
\end{equation}
Set the initial hidden state to zero vector, then we have,
\begin{equation}
    h_1 = x'_1 = \sum_{i=1}^n a^i_1 u_i 
\end{equation}
Compute $h_t$ recursively,
\begin{equation}
    h_t = \sum_{i=1}^n \sum_{j=1}^t a^i_j \lambda_i ^ {j-1} u_i
\label{eq:12}
\end{equation}
Recall (\ref{eq:h_t}), then we have,
\begin{equation}
    \alpha^i_t = \sum_{j=1}^t a^i_j \lambda_i ^ {j-1}
\end{equation}
\begin{equation}
    \alpha^i_t = \lambda_i  \alpha^i_{t-1} + a^i_t
\end{equation}
Therefore, a linear state space model equals a set of first-order IIR filters in complex space along the direction of eigenvectors.

In nonlinear recurrent systems, for the weight matrix of the hidden state, we can still use the same analysis of eigendecomposition as above. However, equation (\ref{eq:12}) breaks due to the nonlinear activation function, causing it difficult to analyze nonlinear cases.

\section{Conjectures for Nonlinear Case}
The activation functions are used to introduce non-linearity in recurrent neural networks. Here, we can rethink it from the view of eigenvalues and eigenvectors of the weight matrix in equation (\ref{eq:decay}). However, the issue is that the information decay of a hidden state is determined by the eigenvalues. A system with small eigenvalues loses information very fast. And in a stable linear system, the eigenvalue norms are strictly limited by 1, which provides no long-term dependency. However, using nonlinear activation functions restricts the outputs either from 0 to 1 ($sigmoid$) or from -1 to 1 ($tanh$). It prevents the hidden states from exploding when the eigenvalues are greater than 1. As for the $ReLU$ activation, there are cases where eigenvalues are negative numbers, meaning the hidden states flip in the opposite direction. And $ReLU$ prevents the exploding by directly cutting the negative parts. Hereby, we make the following conjecture:
\newtheorem{conjecture}{Conjecture}
\begin{conjecture}
    A nonlinear recurrent system improves the long-term dependency by increasing the eigenvalues of the weighting matrix ($> 1$). The larger the eigenvalues are, the better long-term dependency is provided.
\end{conjecture}

To test this conjecture, we perform two experiments on linear-RNN, tanh-RNN, and relu-RNN and compare the eigenvalue norms. We first train the three models on a regression task using the Mackey glass dataset. We use the MSE as the loss function and Adam optimizer with a learning rate of 0.01. We set the size of the hidden state to 8. Since the new hidden state is given by both the previous hidden state and input sample, the eigenvalue norms should be much smaller than 1. From Table \ref{tab:regression} we can see both tanh-RNN and relu-RNN have larger eigenvalue norms than linear-RNN.
\begin{table}[h]
\caption{Eigenvalue Norms on Mackey glass}
\centering
\begin{tabular}{|c|c|c|}
\hline
linear-RNN & tanh-RNN & relu-RNN \\\hline
0.57  & 0.72 & 0.66 \\
0.57  & 0.72 & 0.52\\
0.36  & 0.65 & 0.52\\
0.36 & 0.52 & 0.50\\
0.35 & 0.52 &  0.50\\
0.35 & 0.34 & 0.25\\
0.23 & 0.27 & 0.25\\
0.07 & 0.20 & 0.04 \\\hline
\end{tabular}
\label{tab:regression}
\end{table}

We also train the three networks on the sequential MNIST classification task, which requires a higher demand for long-term dependency. We use the softmax output and cross-entropy as the loss function. Since models are only trained with fixed length sequences and there is no limitation on the values of the outputs (before the softmax), the restriction for stability in the networks is looser. And the eigenvalue norms can be slightly larger than 1 in linear RNN. Besides, in most classification cases, the models extract features from the input and keep information. Larger eigenvalue norms are expected. We set the size of the hidden state to 150 and use Adam with a learning rate of 0.0001. We only show the largest 10th unique eigenvalue norms (only keep one for conjugate eigenvalues) in Table \ref{tab:class}. 
\begin{table}[h]
\caption{Eigenvalues Norms on MNIST classification}
\centering
\begin{tabular}{|c|c|c|}
\hline
linear-RNN & tanh-RNN & relu-RNN \\\hline
1.02  & 1.71 & 1.35 \\
1.01  & 1.60 & 1.31\\
0.99  & 1.53 & 1.30\\
0.98 & 1.41 & 1.29\\
0.98 & 1.27 &  0.98\\
0.96 & 1.19 & 0.88\\
0.95 & 0.95 & 0.81\\
0.94 & 0.90 & 0.74\\
0.91 & 0.89 & 0.70\\
0.83 & 0.74 & 0.68 \\\hline
\end{tabular}
\label{tab:class}
\end{table}

In a stable linear system, the eigenvalue norms cannot be larger than 1, meaning that the information can hardly be preserved in the hidden states. Therefore, the hidden states tend to shrink to zero when propagating forward through time. And the hidden states have a higher density distribution around the origin. It causes discriminating information much more difficult. However, with the nonlinear activation functions, the eigenvalue norms can be greater than 1. Here the space of hidden states is a hypercube limited by the activation functions. These eigenvalues keep pushing the hidden states to the surface of the hypercube. Therefore, the hidden states can explore more space in the hypercube and have a better distribution for discriminant features. In conclusion, we have the following conjecture:
\begin{conjecture}
    For a recurrent system with long-term dependency, the hidden states are capable of exploring the space away from the origin and avoiding collapsing. 
\end{conjecture}

\begin{figure}[h]
    \centering
    \includegraphics[scale=0.5]{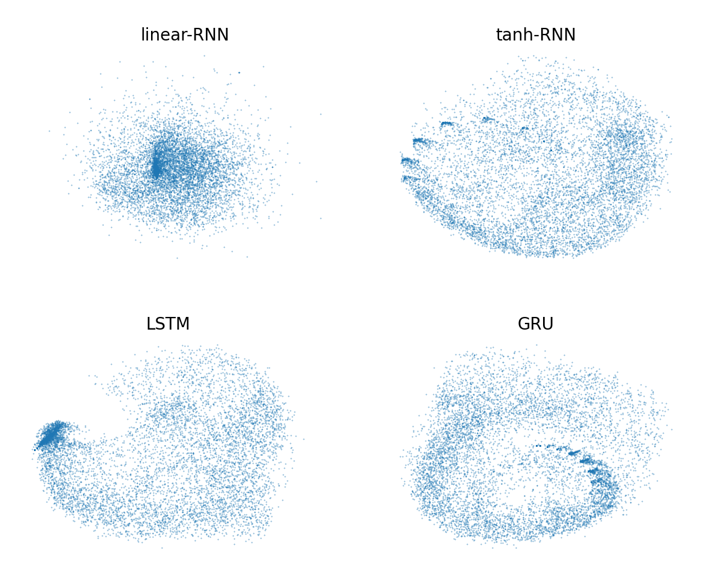}
    \caption{Distribution of Hidden States}
    \label{fig:distribution}
\end{figure}

From the tables above, we notice that the classification task has much better long-term dependency than regression. Therefore, we validate this conjecture using the MNIST dataset and compare the hidden states of linear-RNN,  tanh-RNN, LSTM and GRU. We first train the four networks until the  best performance is achieved. Then, we collect all the hidden states along the trajectories. For visualization, we use the principal component analysis (PCA) and plot the scatters on the first two principal components (Figure \ref{fig:distribution}). We can see that the plots are consistent with our conjecture. Besides, LSTM and GRU show special structures in the distribution which requires future explanation. 

\section{Eigen Initializer}
In gradient descent methods, a proper initialization for the weights not only helps with the convergence but also provides better local optimal solutions. To study the relation between eigenvalues and different initialization methods, we calculate the eigenvalue norms for different initializers (Table \ref{tab:ini}). We create a weighting matrix with a size $8\times8$ and compare the default uniform initializer (in Pytorch) with sp-RNN, Xavier initializer, and kaiming initializer. We generate each initializer 500 times and calculate the ordered mean values. The default initializer of Pytorch is initialized by a uniform distribution with a standard deviation given by,
\begin{equation}
    stdv = 1 / sqrt(n)
\end{equation}
where $n$ is the size of input features.

\begin{table}[h]
\caption{Eigenvalue Norms for different initializers}
\centering
\begin{adjustbox}{width=0.46\textwidth}
\begin{tabular}{|c|c|c|c|c|c|}
\hline
Default & sp-RNN & \begin{tabular}[c]{@{}c@{}}Xavier\\ normal\end{tabular} &\begin{tabular}[c]{@{}c@{}}Xavier\\uniform \end{tabular} & \begin{tabular}[c]{@{}c@{}}kaiming\\ normal\end{tabular} & \begin{tabular}[c]{@{}c@{}}kaiming\\ uniform\end{tabular} \\\hline
0.61$\pm$0.09 & 1.00$\pm$0.00 & 1.06$\pm$0.18 & 1.07$\pm$0.16  & 1.51$\pm$0.26 & 1.51$\pm$0.23\\
0.54$\pm$0.07 & 0.74$\pm$0.12 & 0.95$\pm$0.16 & 0.96$\pm$0.13  & 1.33$\pm$0.22 & 1.35$\pm$0.20\\
0.48$\pm$0.07 & 0.66$\pm$0.10 & 0.83$\pm$0.14 & 0.84$\pm$0.12  & 1.17$\pm$0.19 & 1.19$\pm$0.17\\
0.42$\pm$0.07 & 0.47$\pm$0.08 & 0.73$\pm$0.13 & 0.75$\pm$0.12  & 1.04$\pm$0.19 & 1.06$\pm$0.18\\
0.37$\pm$0.07 & 0.42$\pm$0.06 & 0.63$\pm$0.14 & 0.66$\pm$0.12  & 0.90$\pm$0.18 & 0.92$\pm$0.19\\
0.31$\pm$0.08 & 0.31$\pm$0.05 & 0.52$\pm$0.14 & 0.54$\pm$0.14  & 0.74$\pm$0.19 & 0.77$\pm$0.20\\
0.24$\pm$0.09 & 0.31$\pm$0.05 & 0.39$\pm$0.15 & 0.42$\pm$0.15  & 0.56$\pm$0.19 & 0.59$\pm$0.21\\
0.14$\pm$0.09 & 0.30$\pm$0.04 & 0.24$\pm$0.16 & 0.24$\pm$0.15  & 0.34$\pm$0.22 & 0.35$\pm$0.21\\\hline
\end{tabular}
\end{adjustbox}
\label{tab:ini}
\end{table}

From table \ref{tab:ini}, we notice that compared with other initializers, the default initializer provides small eigenvalues. At the beginning of training, if the eigenvalue norms are small, the information of input samples decays very fast. And the model can hardly learn from previous samples. Other initializers provide larger eigenvalues, helping the model to learn from more samples. From the perspective of eigenvalues, we can also explain why other initializers help with the convergence rate.

Therefore, we can help models learn better long-term dependency with a better initialization on eigenvalues. First, we set all the eigenvalue norms to $\lambda\in(0, 1)$. We want the model can learn more from long-term dependency and let the hidden states store more information in the beginning. Therefore, we set $\lambda$ to 0.95 for all experiments.
\begin{equation}
    W_0 = diag(\lambda)
\end{equation}
where $w_1$ is a diagonal matrix with the size of $n\times n$, and $n$ is the size of hidden states.

In hidden state space, the transition function cannot only magnify the hidden states but should also rotate the hidden states by a certain angle along several directions. Therefore, we mimic this rotation by decomposing the process into $n-1$ step. And at step $i$, we randomly sample an angle $\theta_i$ from a uniform distribution $[0, 2\pi]$. Then, we perform the rotation in the space of $i$th and ($i+1$)th dimension. 
\begin{equation}
    W_i = \left[
    \begin{tabular}{cccccc}
        1 & \multicolumn{4}{c}{$\cdots$} & 0\\
        \multirow{4}{*}{$\vdots$} & $\ddots$ & &  &  & \multirow{4}{*}{$\vdots$}\\
         &  &$\cos{\theta_i}$ & $-\sin{\theta_i}$  &  & \\
         &  &$\sin{\theta_i}$ & $\cos{\theta_i}$ & &  \\
         &  &  &  & $\ddots$ & \\
        0 & \multicolumn{4}{c}{$\cdots$} & 1
    \end{tabular}
    \right]
\end{equation}

Then, the final initializer is,
\begin{equation}
    W = \prod_{i=0} ^{n-1} W_i
\end{equation}

\section{Experiments and Results}
\subsection{Tomita Grammar}
\label{subsec:1}
The Tomita Grammar \cite{tomita:cogsci82} is a set of strings with alphabet $\{0, 1\}$ following predefined rules. It contains 7 different languages and has been widely studied for deterministic finite automata (DFA) extraction. Here, we use the Tomita Grammar to test our initialization method. We compare the learning curves on tanh-RNN, LSTM and GRU with uniform initialization (Pytorch), Xavier initializer and kaiming initializer. We also compare IRNN, sp-RNN for tanh-RNN, which are both RNN-only initializers. We show the comparison of learning curves on grammar 4. To show stability, we train each method 20 times and plot the average learning curve. We use Adam as the optimizer and uniformly set the learning rate to 0.001 with no weighting decay.

\begin{figure}
    \begin{minipage}[b]{1.0\linewidth}
      \centering
      \centerline{\includegraphics[scale=0.3]{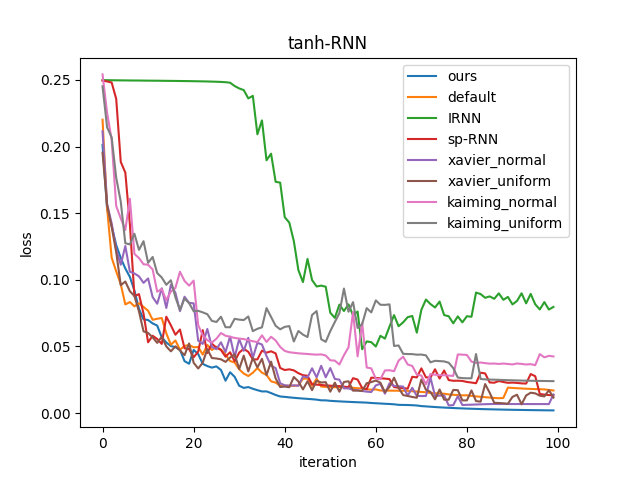}}
    \end{minipage}
    
    \begin{minipage}[b]{1.0\linewidth}
      \centering
      \centerline{\includegraphics[scale=0.3]{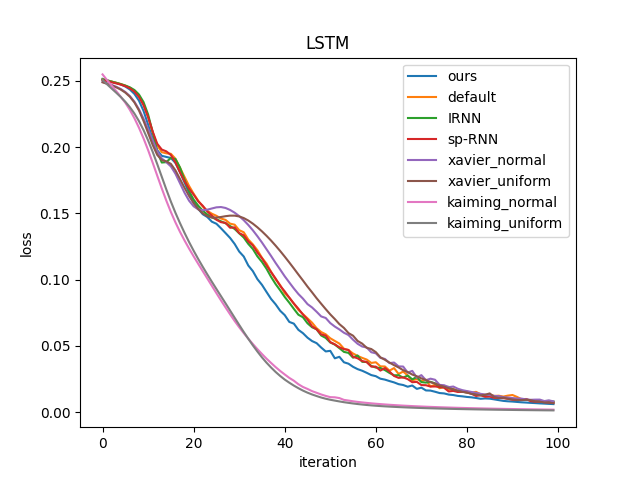}}
    \end{minipage}
    \begin{minipage}[b]{1.0\linewidth}
      \centering
      \centerline{\includegraphics[scale=0.3]{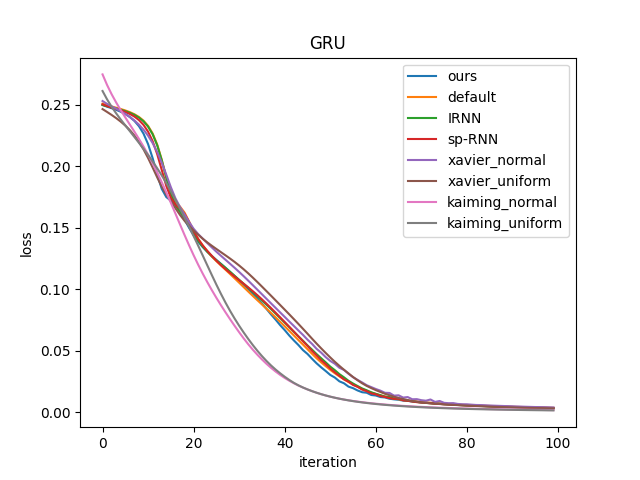}}
    \end{minipage}
    
    \caption{Average Learning Curves on Tomita Grammar 4}
    \label{fig:tomita}
\end{figure}

Figure \ref{fig:tomita} shows that in tanh-RNN, our initialization method provides a smoother learning curve. This is because our initialization method provides a better long-term dependency at the beginning and networks can benefit from this initialization. And in LSTM and GRU, our method shows slight improvement compared with other methods. Note that the Kaiming method takes advantage of the nonsaturating nonlinearity (ReLU) and is therefore not comparable to the others. 

\subsection{MNIST Classification}
\begin{figure}
    \begin{minipage}[b]{1.0\linewidth}
      \centering
      \centerline{\includegraphics[scale=0.3]{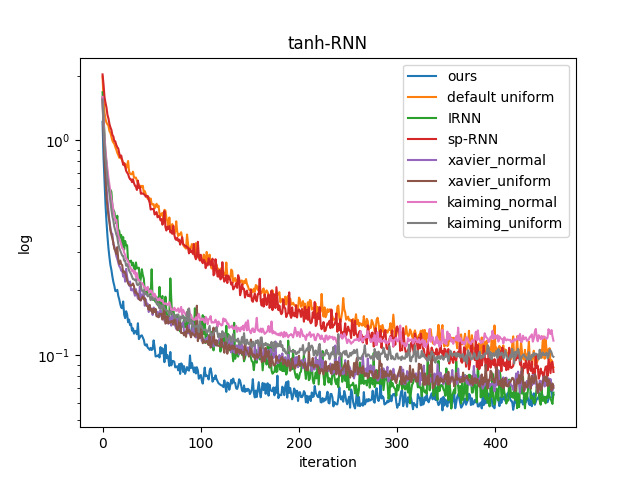}}
    \end{minipage}
    
    \begin{minipage}[b]{1.0\linewidth}
      \centering
      \centerline{\includegraphics[scale=0.3]{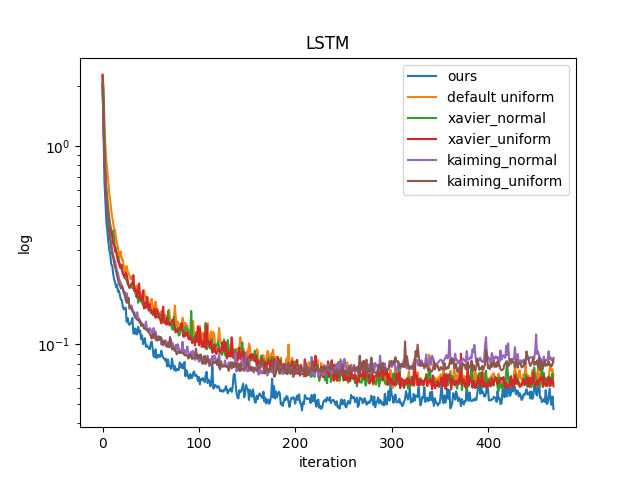}}
    \end{minipage}
    \begin{minipage}[b]{1.0\linewidth}
      \centering
      \centerline{\includegraphics[scale=0.3]{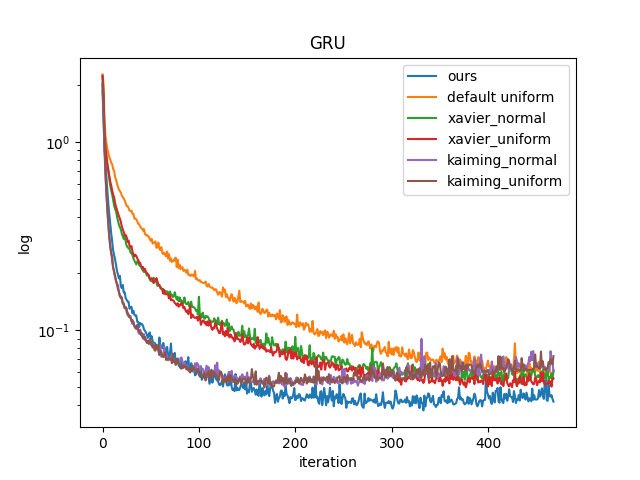}}
    \end{minipage}
    
    \caption{Learning Curves on scanline MNIST}
    \label{fig:sMNIST}
\end{figure}

We also test our initialization method on MNIST classification. In the experiment, we use the sequential MNIST by scanlines. Each time 28 pixels from the same row are presented as the input and there are 28 inputs in total. This is a simple task and we only care about the convergence of the learning curves. We set the same comparison as in section \ref{subsec:1}. We set the hidden states for all networks to 150 and use cross-entropy as the loss function. We use Adam as the optimizer with a learning rate of 0.0001. The result is shown in Figure \ref{fig:sMNIST}. Our method provides both a good convergence rate and better local optima for all networks. Still, we only have a significant improvement on the convergence rate on tanh-RNN. But in LSTM and GRU, we also benefit from learning the long-term dependency.

\subsection{Multi30k Machine Translation}
The Multi30k \cite{DBLP:journals/corr/ElliottFSS16} is a multi-model dataset that has been widely used in machine translation and image description. Each sample contains an image and a pair of descriptions in German and English. In this paper, we only take the sentences and perform a machine translation task from German to English.
\begin{figure}
    \begin{minipage}[b]{1.0\linewidth}
      \centering
      \centerline{\includegraphics[scale=0.3]{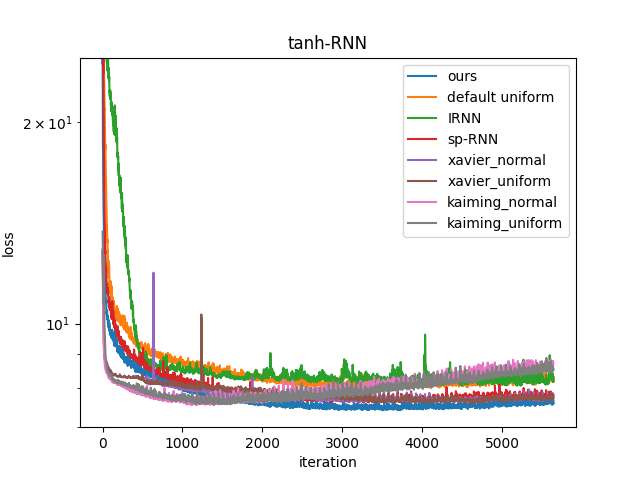}}
    \end{minipage}
    
    \begin{minipage}[b]{1.0\linewidth}
      \centering
      \centerline{\includegraphics[scale=0.3]{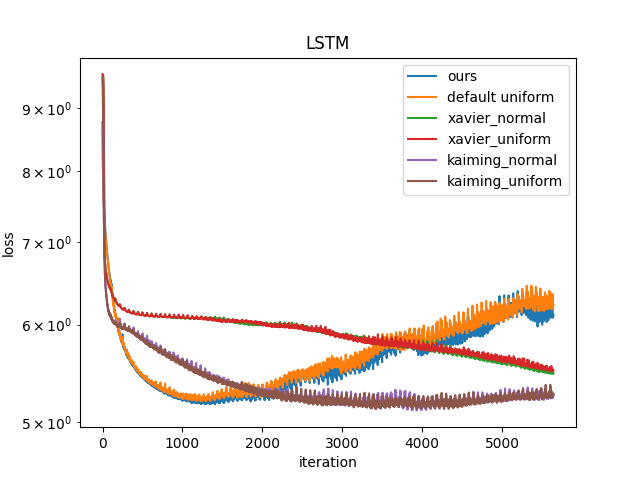}}
    \end{minipage}
    \begin{minipage}[b]{1.0\linewidth}
      \centering
      \centerline{\includegraphics[scale=0.3]{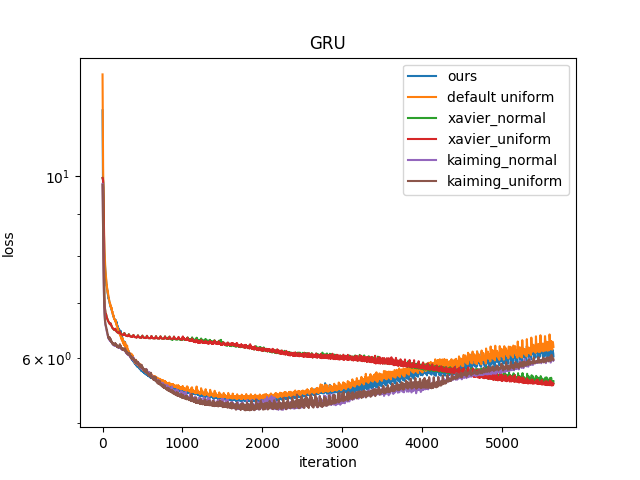}}
    \end{minipage}
    
    \caption{Learning Curves on Multi30k}
    \label{fig:multi30k}
\end{figure}
In the experiment, we choose the encoder-decoder architecture for sequence-to-sequence prediction. Each encoder and decoder contains an embedding layer and a recurrent layer (either tanh-RNN, LSTM or GRU). For simplicity, we only use one layer of each recurrent layer. For data prepossessing, we use spaCy to extract tokens from the sentences, which is an industrial package for natural language processing. Then, we replace all tokens that appear only once with the unknown token [UNK]. We also use the [PAD] token for padding the sequences. We uniformly set the embedding size to 512 and the hidden size to 256. We compare our methods with other initialization methods on tanh-RNN, LSTM and GRU. We use the cross-entropy as the loss function with a label smooth of 0.1. We also use Adam as the optimizer with a learning rate of 0.1. The learning curves are shown in Figure \ref{fig:multi30k}. We overtrain the LSTM and GRU cases to display the full convergence of the other methods.

In the tanh-RNN, our method outperforms default uniform, IRNN and sp-RNN. It also provides a better and more stable solution compared with all other initialization methods. In LSTM, our methods provide the best convergence rate as the default uniform initialization and in GRU, our performance is pretty close to all other methods, except Xavier initializers fail to provide a good convergence.  

\section{Conclusion}
In this paper, we analyze the dynamics of hidden states in recurrent systems. We study the relationship between eigenvalues and long-term dependency in linear recurrent models and testify to the conjectures for nonlinear systems. We also propose a new eigen initializer for all kinds of recurrent neural networks. We test is on multiple datasets. Our results show that our initialization methods can not only provide a good convergence rate but can also provide a better local optimal solution. In some experiments, we outperform Xavier initializer and kaiming initializer. We provide a new alternative initializer for training recurrent neural networks and an intuitive explanation of why it works.

\bibliographystyle{IEEEtranS}
\bibliography{ref}

\begin{thebibliography}{10}
\providecommand{\url}[1]{#1}
\csname url@samestyle\endcsname
\providecommand{\newblock}{\relax}
\providecommand{\bibinfo}[2]{#2}
\providecommand{\BIBentrySTDinterwordspacing}{\spaceskip=0pt\relax}
\providecommand{\BIBentryALTinterwordstretchfactor}{4}
\providecommand{\BIBentryALTinterwordspacing}{\spaceskip=\fontdimen2\font plus
\BIBentryALTinterwordstretchfactor\fontdimen3\font minus
  \fontdimen4\font\relax}
\providecommand{\BIBforeignlanguage}[2]{{%
\expandafter\ifx\csname l@#1\endcsname\relax
\typeout{** WARNING: IEEEtranS.bst: No hyphenation pattern has been}%
\typeout{** loaded for the language `#1'. Using the pattern for}%
\typeout{** the default language instead.}%
\else
\language=\csname l@#1\endcsname
\fi
#2}}
\providecommand{\BIBdecl}{\relax}
\BIBdecl

\bibitem{allen2019convergence}
Z.~Allen-Zhu, Y.~Li, and Z.~Song, ``On the convergence rate of training
  recurrent neural networks,'' \emph{Advances in neural information processing
  systems}, vol.~32, 2019.

\bibitem{arjovsky2016unitary}
M.~Arjovsky, A.~Shah, and Y.~Bengio, ``Unitary evolution recurrent neural
  networks,'' in \emph{International conference on machine learning}.\hskip 1em
  plus 0.5em minus 0.4em\relax PMLR, 2016, pp. 1120--1128.

\bibitem{chang2019antisymmetricrnn}
B.~Chang, M.~Chen, E.~Haber, and E.~H. Chi, ``Antisymmetricrnn: A dynamical
  system view on recurrent neural networks,'' \emph{arXiv preprint
  arXiv:1902.09689}, 2019.

\bibitem{GRU}
\BIBentryALTinterwordspacing
K.~Cho, B.~van Merrienboer, {\c{C}}.~G{\"{u}}l{\c{c}}ehre, D.~Bahdanau,
  F.~Bougares, H.~Schwenk, and Y.~Bengio, ``Learning phrase representations
  using {RNN} encoder-decoder for statistical machine translation,'' in
  \emph{Proceedings of the 2014 Conference on Empirical Methods in Natural
  Language Processing, {EMNLP} 2014, October 25-29, 2014, Doha, Qatar, {A}
  meeting of SIGDAT, a Special Interest Group of the {ACL}}, A.~Moschitti,
  B.~Pang, and W.~Daelemans, Eds.\hskip 1em plus 0.5em minus 0.4em\relax {ACL},
  2014, pp. 1724--1734. [Online]. Available:
  \url{https://doi.org/10.3115/v1/d14-1179}
\BIBentrySTDinterwordspacing

\bibitem{DBLP:journals/corr/ElliottFSS16}
\BIBentryALTinterwordspacing
D.~Elliott, S.~Frank, K.~Sima'an, and L.~Specia, ``Multi30k: Multilingual
  english-german image descriptions,'' \emph{CoRR}, vol. abs/1605.00459, 2016.
  [Online]. Available: \url{http://arxiv.org/abs/1605.00459}
\BIBentrySTDinterwordspacing

\bibitem{glorot2010understanding}
X.~Glorot and Y.~Bengio, ``Understanding the difficulty of training deep
  feedforward neural networks,'' in \emph{Proceedings of the thirteenth
  international conference on artificial intelligence and statistics}.\hskip
  1em plus 0.5em minus 0.4em\relax JMLR Workshop and Conference Proceedings,
  2010, pp. 249--256.

\bibitem{NTM}
A.~Graves, G.~Wayne, and I.~Danihelka, ``Neural turing machines,'' \emph{arXiv
  preprint arXiv:1410.5401}, 2014.

\bibitem{DNC}
A.~Graves, G.~Wayne, M.~Reynolds, T.~Harley, I.~Danihelka,
  A.~Grabska-Barwi{\'n}ska, S.~G. Colmenarejo, E.~Grefenstette, T.~Ramalho,
  J.~Agapiou \emph{et~al.}, ``Hybrid computing using a neural network with
  dynamic external memory,'' \emph{Nature}, vol. 538, no. 7626, pp. 471--476,
  2016.

\bibitem{NeuralStack}
E.~Grefenstette, K.~M. Hermann, M.~Suleyman, and P.~Blunsom, ``Learning to
  transduce with unbounded memory,'' \emph{Advances in neural information
  processing systems}, vol.~28, 2015.

\bibitem{gu2020improving}
A.~Gu, C.~Gulcehre, T.~Paine, M.~Hoffman, and R.~Pascanu, ``Improving the
  gating mechanism of recurrent neural networks,'' in \emph{International
  Conference on Machine Learning}.\hskip 1em plus 0.5em minus 0.4em\relax PMLR,
  2020, pp. 3800--3809.

\bibitem{he2015delving}
K.~He, X.~Zhang, S.~Ren, and J.~Sun, ``Delving deep into rectifiers: Surpassing
  human-level performance on imagenet classification,'' in \emph{Proceedings of
  the IEEE international conference on computer vision}, 2015, pp. 1026--1034.

\bibitem{hochreiter1997long}
S.~Hochreiter and J.~Schmidhuber, ``Long short-term memory,'' \emph{Neural
  computation}, vol.~9, no.~8, pp. 1735--1780, 1997.

\bibitem{jaeger2002adaptive}
H.~Jaeger, ``Adaptive nonlinear system identification with echo state
  networks,'' \emph{Advances in neural information processing systems},
  vol.~15, 2002.

\bibitem{gatebias}
\BIBentryALTinterwordspacing
R.~Jozefowicz, W.~Zaremba, and I.~Sutskever, ``An empirical exploration of
  recurrent network architectures,'' in \emph{Proceedings of the 32nd
  International Conference on Machine Learning}, ser. Proceedings of Machine
  Learning Research, F.~Bach and D.~Blei, Eds., vol.~37.\hskip 1em plus 0.5em
  minus 0.4em\relax Lille, France: PMLR, 07--09 Jul 2015, pp. 2342--2350.
  [Online]. Available:
  \url{https://proceedings.mlr.press/v37/jozefowicz15.html}
\BIBentrySTDinterwordspacing

\bibitem{DMN}
A.~Kumar, O.~Irsoy, P.~Ondruska, M.~Iyyer, J.~Bradbury, I.~Gulrajani, V.~Zhong,
  R.~Paulus, and R.~Socher, ``Ask me anything: Dynamic memory networks for
  natural language processing,'' in \emph{International conference on machine
  learning}.\hskip 1em plus 0.5em minus 0.4em\relax PMLR, 2016, pp. 1378--1387.

\bibitem{le2015simple}
Q.~V. Le, N.~Jaitly, and G.~E. Hinton, ``A simple way to initialize recurrent
  networks of rectified linear units,'' \emph{arXiv preprint arXiv:1504.00941},
  2015.

\bibitem{lim2021noisy}
S.~H. Lim, N.~B. Erichson, L.~Hodgkinson, and M.~W. Mahoney, ``Noisy recurrent
  neural networks,'' \emph{Advances in Neural Information Processing Systems},
  vol.~34, pp. 5124--5137, 2021.

\bibitem{ma2020temporal}
Q.~Ma, Z.~Lin, E.~Chen, and G.~Cottrell, ``Temporal pyramid recurrent neural
  network,'' in \emph{Proceedings of the AAAI Conference on Artificial
  Intelligence}, vol.~34, no.~04, 2020, pp. 5061--5068.

\bibitem{ma2020particle}
X.~Ma, P.~Karkus, D.~Hsu, and W.~S. Lee, ``Particle filter recurrent neural
  networks,'' in \emph{Proceedings of the AAAI Conference on Artificial
  Intelligence}, vol.~34, no.~04, 2020, pp. 5101--5108.

\bibitem{massart2022coordinate}
E.~Massart and V.~Abrol, ``Coordinate descent on the orthogonal group for
  recurrent neural network training,'' in \emph{Proceedings of the AAAI
  Conference on Artificial Intelligence}, vol.~36, no.~7, 2022, pp. 7744--7751.

\bibitem{RNNEM}
B.~Peng, K.~Yao, L.~Jing, and K.-F. Wong, ``Recurrent neural networks with
  external memory for spoken language understanding,'' in \emph{Natural
  Language Processing and Chinese Computing}, J.~Li, H.~Ji, D.~Zhao, and
  Y.~Feng, Eds.\hskip 1em plus 0.5em minus 0.4em\relax Cham: Springer
  International Publishing, 2015, pp. 25--35.

\bibitem{rotman2021shuffling}
M.~Rotman and L.~Wolf, ``Shuffling recurrent neural networks,'' in
  \emph{Proceedings of the AAAI Conference on Artificial Intelligence},
  vol.~35, no.~11, 2021, pp. 9428--9435.

\bibitem{MemN2N}
S.~Sukhbaatar, J.~Weston, R.~Fergus \emph{et~al.}, ``End-to-end memory
  networks,'' \emph{Advances in neural information processing systems},
  vol.~28, 2015.

\bibitem{talathi2015improving}
S.~S. Talathi and A.~Vartak, ``Improving performance of recurrent neural
  network with relu nonlinearity,'' \emph{arXiv preprint arXiv:1511.03771},
  2015.

\bibitem{tomita:cogsci82}
M.~Tomita, ``Dynamic construction of finite automata from examples using
  hill-climbing,'' in \emph{{P}roceedings of the Fourth Annual Conference of
  the Cognitive Science Society}, Ann Arbor, Michigan, 1982, pp. 105--108.

\bibitem{stateregularized}
C.~Wang and M.~Niepert, ``State-regularized recurrent neural networks,'' in
  \emph{International Conference on Machine Learning}.\hskip 1em plus 0.5em
  minus 0.4em\relax PMLR, 2019, pp. 6596--6606.

\bibitem{whitney1984kalman}
D.~WHITNEY, ``The kalman filter family tree- a survey of state-of-the-art
  analysis methods based on the kalman filter,'' \emph{NAECON 1984}, pp.
  426--432, 1984.

\bibitem{zhang2021sbo}
Z.~Zhang, Y.~Yue, G.~Wu, Y.~Li, and H.~Zhang, ``Sbo-rnn: Reformulating
  recurrent neural networks via stochastic bilevel optimization,''
  \emph{Advances in Neural Information Processing Systems}, vol.~34, pp.
  25\,839--25\,851, 2021.

\end{thebibliography}
\end{document}